\documentclass[letterpaper]{article} % DO NOT CHANGE THIS
\pdfoutput=1
\usepackage[submission]{aaai25}  % DO NOT CHANGE THIS
\usepackage{times}  % DO NOT CHANGE THIS
\usepackage{helvet}  % DO NOT CHANGE THIS
\usepackage{courier}  % DO NOT CHANGE THIS
\usepackage[hyphens]{url}  % DO NOT CHANGE THIS
\usepackage{graphicx} % DO NOT CHANGE THIS
\usepackage{natbib}  % DO NOT CHANGE THIS AND DO NOT ADD ANY OPTIONS TO IT
\usepackage{caption} % DO NOT CHANGE THIS AND DO NOT ADD ANY OPTIONS TO IT
\usepackage{algorithm, algorithmic}
\usepackage{amsmath,amssymb,amsfonts}
\usepackage{color}
\usepackage{multirow,booktabs}
\usepackage{makecell}
\usepackage{newfloat}
\usepackage{listings}
\DeclareCaptionStyle{ruled}{labelfont=normalfont,labelsep=colon,strut=off} % DO NOT CHANGE THIS
\floatstyle{ruled}
\newfloat{listing}{tb}{lst}{}
\floatname{listing}{Listing}
\title{PanAdapter: Two-Stage Fine-Tuning with Spatial-Spectral Priors Injecting for Pansharpening}

\author{
    RuoCheng Wu\textsuperscript{\rm 1}\equalcontrib,
    ZiEn Zhang\textsuperscript{\rm 1}\equalcontrib,
    ShangQi Deng\textsuperscript{\rm 1},
    YuLe Duan\textsuperscript{\rm 1},
    LiangJian Deng\textsuperscript{\rm 1}
}
\affiliations{
    \textsuperscript{\rm 1}University of Electronic Science and Technology of China\\
    fantastiquebookah@gmail.com, zienzhang314@gmail.com, shangqideng0124@gmail, duanyll@std.uestc.edu.cn, liangjian.deng@uestc.edu.cn
}

\usepackage{bibentry}
\begin{document}

\maketitle

\begin{abstract}
Pansharpening is a challenging image fusion task that involves restoring images using two different modalities: low-resolution multispectral images (LRMS) and high-resolution panchromatic (PAN). Many end-to-end specialized models based on deep learning (DL) have been proposed, yet the scale and performance of these models are limited by the size of dataset. Given the superior parameter scales and feature representations of pre-trained models, they exhibit outstanding performance when transferred to downstream tasks with small datasets. Therefore, we propose an efficient fine-tuning method, namely PanAdapter, which utilizes additional advanced semantic information from pre-trained models to alleviate the issue of small-scale datasets in pansharpening tasks. Specifically, targeting the large domain discrepancy between image restoration and pansharpening tasks, the PanAdapter adopts a two-stage training strategy for progressively adapting to the downstream task. In the first stage, we fine-tune the pre-trained CNN model and extract task-specific priors at two scales by proposed Local Prior Extraction (LPE) module. In the second stage, we feed the extracted two-scale priors into two branches of cascaded adapters respectively. At each adapter, we design two parameter-efficient modules for allowing the two branches to interact and be injected into the frozen pre-trained VisionTransformer (ViT) blocks. We demonstrate that by only training the proposed LPE modules and adapters with a small number of parameters, our approach can benefit from pre-trained image restoration models and achieve state-of-the-art performance in several benchmark pansharpening datasets. The code will be available soon.
\end{abstract}


\begin{figure}[!ht]
    \scriptsize
    \setlength{\tabcolsep}{0.7pt}
    \centering
    \includegraphics[width=0.48\textwidth]{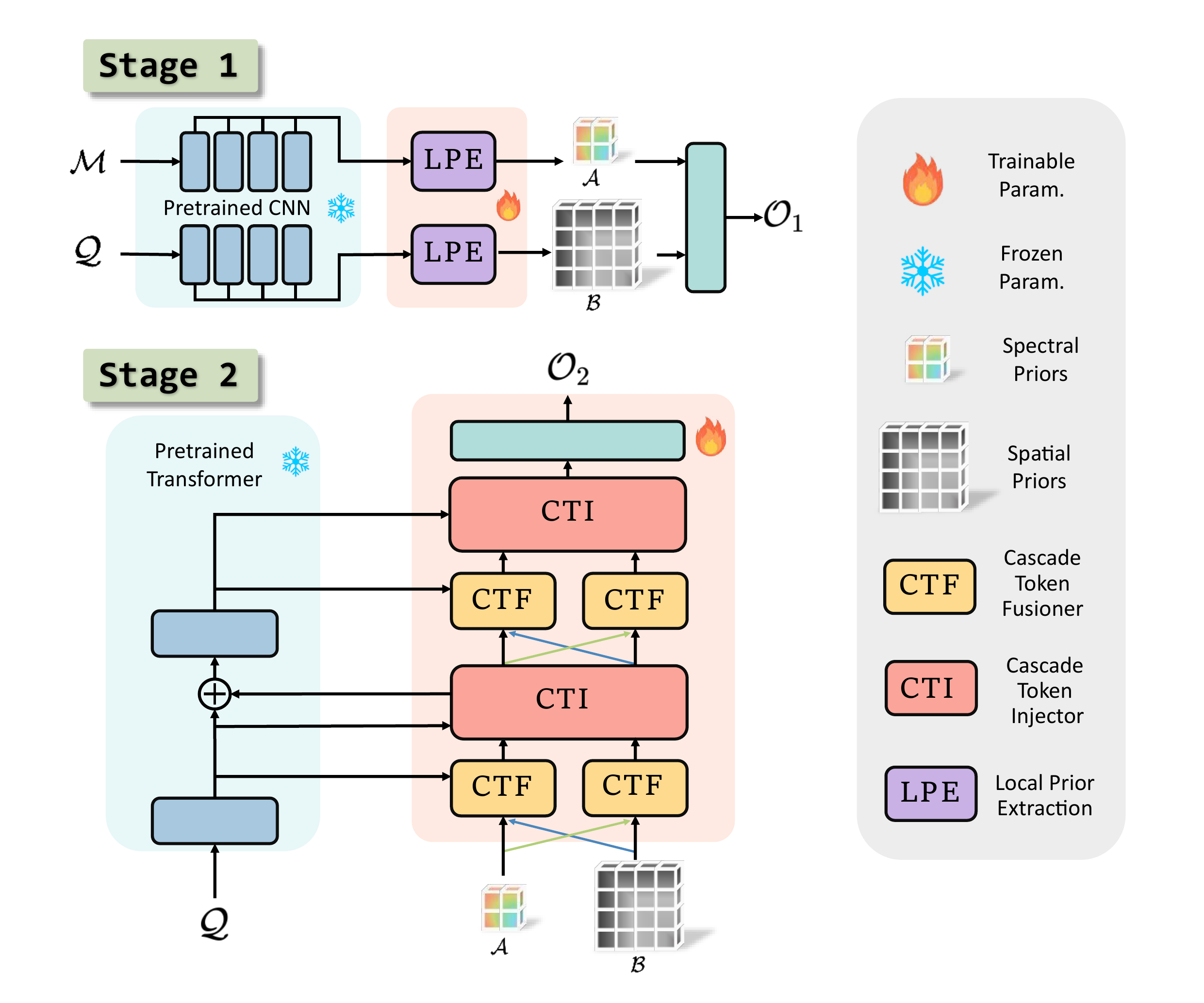}
    \caption{PanAdapter's two-stage fine-tuning framework.} 
    \label{network}
\end{figure}

\section{Introduction}

Due to the hardware limitations, existing sensors only acquire low-resolution multispectral images (LRMS) and high-resolution panchromatic (PAN) images. Pansharpening fuses LRMS and PAN images to produce high-resolution multispectral images (HRMS). Various pansharpening methods have been proposed, including traditional and deep learning-based methods. Traditional methods include component substitution (CS) techniques~\cite{choi2010new, vivone2019robust}, multi-resolution analysis (MRA) methods~\cite{vivone2013contrast, vivone2018full}, and variational optimization-based (VO) approaches~\cite{palsson2013new, he2014new, yan2022panchromatic, zhou2022normalization}. In recent years, convolutional neural networks (CNN) have been widely applied to pansharpening~\cite{masi2016pansharpening}. Benefiting from the inductive bias and residual structures of CNNs, increasingly deeper networks have been proposed to address pansharpening tasks, such as PanNet~\cite{yang2017pannet}, MSDCNN~\cite{wei2017multi}, BDPN~\cite{zhang2019pan}, DiCNN~\cite{he2019pansharpening}, FusionNet~\cite{deng2020detail}, MSDRN~\cite{wang2021msdrn} and PMACNet~\cite{liang2022pmacnet}. However, because CNNs struggle to extract global information and high-level semantic features, researchers explore methods based on Transformers. Leveraging long-range dependencies and scalability to pre-trained models and datasets, models such as full Transformers~\cite{zhou2022panformer, meng2022vision} and hybrids of CNNs and Transformers~\cite{zhu2023mutiscale, su2023ctcp} have made progress. Due to the limited size of dataset (500 to 10,000 images), blindly increasing  model parameters is likely to lead to overfitting. To benefit from larger datasets and pre-trained models, we introduce the paradigm of pre-training and fine-tuning into pansharpening. By fine-tuning upstream image restoration models, we aim to enhance the performance of pansharpening tasks.

Recently, the field of image restoration has explored pre-trained models to address various image degradation problems~\cite{chen2021pre, liu2021swin, wang2022uformer}. Since pansharpening essentially is solving an inverse problem of image degradation, pre-trained image restoration models are helpful in addressing the pansharpening task. IPT~\cite{chen2021pre} introduced the first Transformer-based pre-trained model for low-level tasks, followed by approaches like Swin Transformer~\cite{liu2021swin}, and Uformer~\cite{wang2022uformer}. To avoid significant parameter size and high training costs with full fine-tuning, we explore parameter-efficient fine-tuning (PEFT) strategies. In the field of PEFT, the methods of inserting small modules first emerge, such as Adapter~\cite{houlsby2019parameter} and LoRA~\cite{hu2021lora}. Prompt-based approaches~\cite{lester2021power, li2021prefix} add a prefix prompt before the input embeddings. BitFit~\cite{zaken2021bitfit} adjusts only the bias terms of the model, while LST~\cite{sung2022lst} constructs side-tuning networks. In the field of computer vision, AdaptFormer~\cite{chen2022adaptformer} adapts pre-trained ViT models, while ViT-Adapter~\cite{chen2022vision} integrates CNNs to capture local priors and compensate for ViT limitations.

However, due to two primary reasons, existing models struggle to achieve optimal performance on pansharpening tasks. Firstly, there is a significant domain gap, where models trained on natural image datasets face challenges in restoring satellite images. Secondly, conventional image restoration models lack the capability to effectively handle multi-modal inputs or multi-scale images, limiting their ability to capture the intricate spatial and spectral priors required for pansharpening. Experiments in Sec.~\ref{finetune_exp} demonstrate that existing fine-tuning strategies have shown limited performance.

Inspired by the multi-stage training strategy in image restoration domain~\cite{zamir2021multi}, we propose a progressive two-stage training strategy to fine-tune a pre-trained CNN model and a pre-trained ViT model addressing the above-mentioned issues. In the first stage, we fine-tune the pre-trained CNN model. Specifically, we insert the Local Prior Extraction (LPE) module to extract spatial features from PAN and spectral features from LRMS, respectively. In the second stage, we inject the priors acquired from the two branches of the CNN network output into the pre-trained ViT using a set of cascaded adapters. Furthermore, inspired by ViT-Adapter~\cite{chen2022vision}, we propose two parameter-efficient interaction modules in the second stage: the Cascade Token Fusioner (CTF) module and the Cascade Token Injector (CTI) module. Referring to existing side-tuning fine-tuning methods~\cite{sung2022lst}, our adapters adopt a cascaded architecture with intermediate results obtained from the backbone network as supplementary inputs for feature extraction, as shown in Fig.~\ref{network}. The main contributions of this paper are as follows:

\begin{itemize}
    \item[1.] We propose PanAdapter, the first parameter-efficient fine-tuning framework designed specifically for the pansharpening task. Successfully, we apply image restoration models pre-trained on natural images to the remote sensing images. We evaluate our method on the WV3, QB and GF2 datasets, achieving state-of-the-art performance compared to various pansharpening methods.
    \item[2.] We develop a novel two-stage fine-tuning strategy to reduce domain transfer difficulty and address convergence issues in the network. In the Local Prior Extraction Stage, we fine-tune a smaller pre-trained CNN network. In the Multiscale Feature Interaction Stage, we construct a dual-branch structure and fine-tune the pre-trained ViT network based on the output from the first stage.
    \item[3.] We design a set of cascaded dual-branch adapters for fusing spatial and spectral priors from the first stage and injecting them into the pre-trained ViT. Through the two modules proposed in cascaded adapters, multiscale information is retained and interacted with intermediate features of the ViT backbone, which effectively balances the network's ability to extract spatial details and fuse spectral information.
\end{itemize}

\section{Related Works}

    \subsection{DL-Based Pansharpening}

    Early attempts for pansharpening primarily focus on using traditional methods~\cite{aiazzi2002context, palsson2013new, vivone2019robust}. However, these methods are often limited by their ability to extract meaningful features. PNN~\cite{masi2016pansharpening} is one of the earliest works to apply CNNs to pansharpening tasks. To boost performance, more and more researchers are working on creating larger and deeper CNN architectures for pansharpening such as PanNet~\cite{yang2017pannet}, BDPN~\cite{zhang2019pan}, FusionNet~\cite{deng2020detail}. Subsequently, to better extract spatial domain information, researchers propose multiscale approaches, such as PMACNet~\cite{liang2022pmacnet} and BiMPan~\cite{hou2023bidomain}. Some adaptive convolution methods have also shown good performance, addressing the spatial invariance property of traditional CNNs, such as LAGConv~\cite{jin2022lagconv}, CANNet~\cite{duan2024content}.     

    Transformers~\cite{vaswani2017attention} excel in capturing global information and long-range dependencies, which are crucial for pansharpening. PanFormer~\cite{zhou2022panformer} marks the first application of Transformer in the pansharpening domain. Subsequently, researchers construct various structures to learn the global and local feature by combining Transformer and CNN, such as MHATP-Net~\cite{zhu2023mutiscale}, CTCP~\cite{su2023ctcp}. However, due to the small size of the dataset, the existing models hold relatively small parameters, and are difficult to scale up to the level of visual backbones.

    \subsection{PEFT for Vision}

    Parameter-efficient fine-tuning (PEFT) aims to adapt pre-trained large models to new tasks by updating or adding parameters. Due to the similarity between language models and computer vision, large model fine-tuning methods are quickly applied to computer vision tasks. The methods of inserting additional parameters are applied to PEFT firstly, such as Adapters~\cite{houlsby2019parameter}. LoRA~\cite{hu2021lora} adds low-rank matrices to key layers of the model to adjust its behavior. In addition, some methods select a subset of parameters from the pre-trained model for updating without altering the network structure, such as Bitfit~\cite{zaken2021bitfit}. Subsequently, side network approaches emerges, which  builds a ladder on top of the original large model, such as Side-Tuning~\cite{zhang2020side}, LST~\cite{sung2022lst}, where part of the output from the large model's layers serves as the input for the ladder model. Visual adapter~\cite{chen2022adaptformer}, Convpass~\cite{jie2022convolutional}, ViT-adapter~\cite{chen2022vision} employ fine-tuning methods to effectively adapt pre-trained ViT to various image and video tasks.

    \subsection{Motivation}

    Recently, due to the lack of long-distance modeling and high-level semantic information capture capabilities of CNNs, some works have begun to attempt to use Transformers for modeling in the pansharpening task~\cite{zhou2022panformer,meng2022vision}. However, because of the limitation in obtaining remote sensing images, the dataset for the pansharpening task is usually small (typically ranging from 500 to 10,000 images), thus scaling up models in the pansharpening task is extremely challenging. Therefore, a natural idea is to search for pre-trained models and fine-tune on pansharpening tasks. Given the similarity in task characteristics, pre-trained image restoration models~\cite{chen2021pre, liu2021swin} are the preferred candidates. Nevertheless, our experiments indicate that existing fine-tuning methods perform poorly when directly adapting image restoration models to pansharpening tasks (refer to Sec.~\ref{finetune_exp}). To address existing difficulties, we propose a two-stage training framework for parameter-efficient fine-tuning. In the first stage, we fine-tune a small pre-trained CNN model by inserting the proposed Local Priors Extraction (LPE) modules to extract spatial and spectral priors at two different scales. In the second stage, based on the fine-tuned CNN model, we construct cascaded adapters for fine-tuning the pre-trained ViT model. Additionally, as the original ViT network does not contain multi-scale information, we design two branches of different scales for retaining and injecting multi-scale information tailored for pansharpening tasks. To enhance the integration efficiency of two branches, we apply the Implicit Neural Representation paradigm as decoding module at the tail end of the fusion.

\section{Method}

    \subsection{Overall Architecture}   \label{sec:architecture}
    As illustrated in Fig.~\ref{network}, the overall training process of our fine-tuning method consists of two stages.

\begin{figure}[t]
        \scriptsize
        \setlength{\tabcolsep}{0.7pt}
        \centering
        \includegraphics[width=0.45\textwidth]{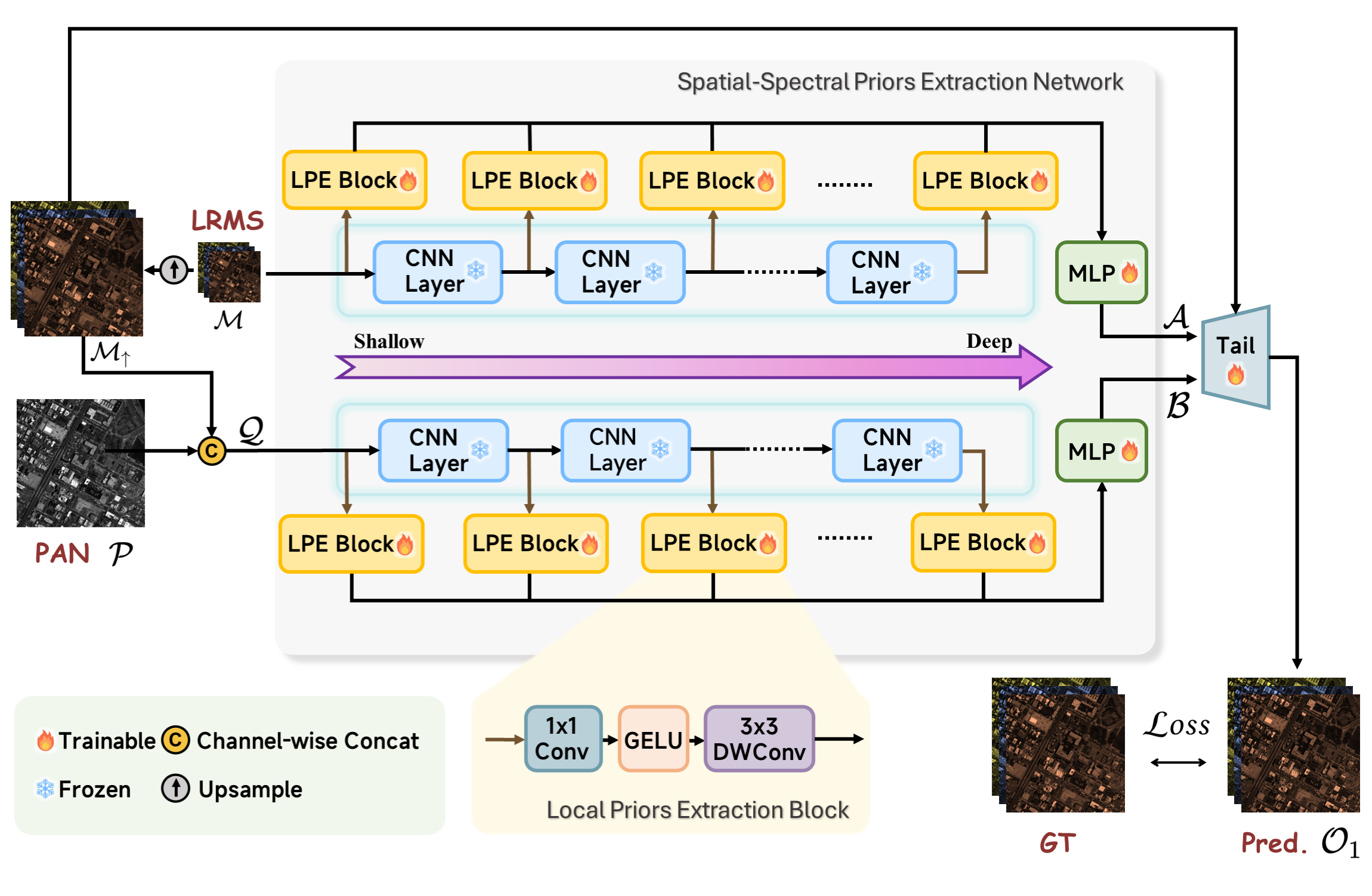}
        \caption{Network structure of the first stage, \emph{i.e.}, the Local Prior Extraction Stage, and the details about the Local Prior Extraction (LPE) block. The frozen CNN networks are pre-trained EDSR~\cite{lim2017enhanced}.}
        \label{stage1}
    \end{figure}

    In the first stage, we fine-tune two pre-trained CNNs trained for super-resolution~\cite{lim2017enhanced} in parallel to obtain two different scales of feature priors. The complete process can be formulated as: 
    $$\mathcal{A}, \mathcal{B} = \text{SSPEN}(\mathcal{Q},\mathcal{M}),$$
    where
    $$\mathcal{Q} = \text{Concat}(\mathcal{M}_\uparrow, \mathcal{P}).$$
    \noindent $\mathcal{M}\in\mathbb{R}^{h\times w\times s}$ denotes the LRMS image, $\mathcal{M}_\uparrow\in\mathbb{R}^{H\times W\times s}$ denotes the upsampled LRMS image and $\mathcal{P}\in\mathbb{R}^{H\times W\times 1}$ denotes the PAN image. In addition, $\text{Concat}(\cdot,\cdot)$ means the concatenation operation in channel dimension. $\text{SSPEN}(\cdot,\cdot)$ means the Spatial-Spectral Priors Extraction Network, denoting the whole pre-trained and fine-tuning network of the first stage, which is illustrated in Fig.~\ref{stage1}. $\mathcal{A}\in\mathbb{R}^{h\times w\times m}$ and $\mathcal{B}\in\mathbb{R}^{H\times W\times m}$ are the local spectral prior and the local spatial prior respectively, representing the results of $\text{SSPEN}$. Here $m$ denotes the feature dimension of the acquired priors. After that, $\mathcal{A}$ and $\mathcal{B}$ are fed into the Tail network for decoding to obtain the predicted image of the first stage $\mathcal{O}_{1}$: 
    $$\mathcal{O}_{1} = \text{Tail}_{1}(\mathcal{A}, \mathcal{B}),$$
    where $\text{Tail}_{1}(\cdot,\cdot)$ denotes the Tail network of the first stage.

\begin{figure*}[t]
        \scriptsize
        \setlength{\tabcolsep}{0.7pt}
        \centering
        \includegraphics[width=16.0cm, height=8.0cm]{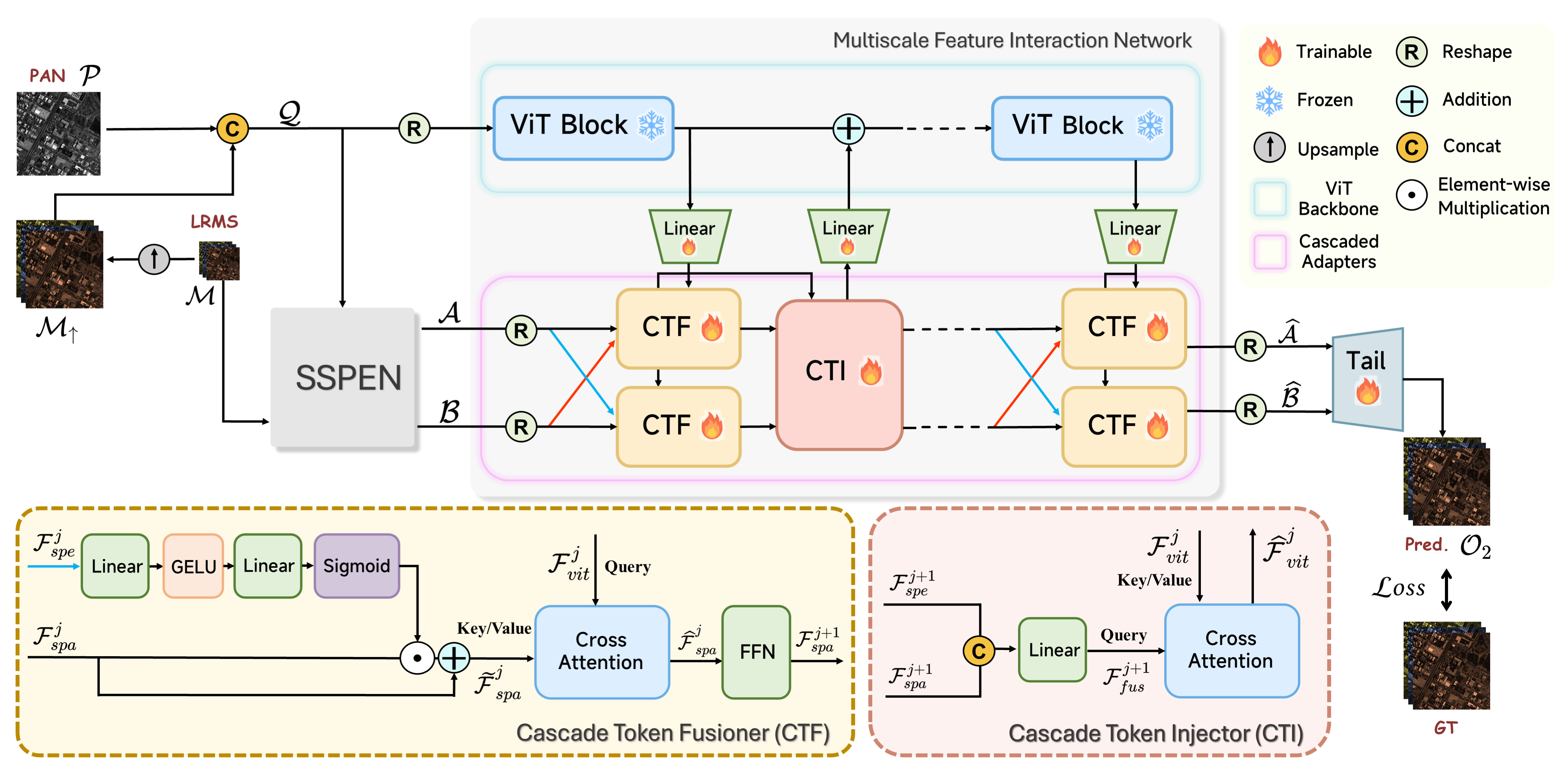}
        \caption{Network structure of the second stage, \emph{i.e.}, the Multiscale Feature Interaction Stage, and the details about the Cascade Token Fusioner (CTF) and the Cascade Token Injector (CTI). The Spatial-Spectral Priors Extraction Network (SSPEN) and the Multiscale Feature Interaction Stage (MFIN) denote the pre-trained and fine-tuning networks of the first stage and the second stage, respectively. The frozen ViT is pre-trained Image Processing Transformer (IPT)~\cite{chen2021pre}.}
        \label{stage2}
    \end{figure*}

    In the second stage, we fine-tune a ViT backbone which is pre-trained for image restoration. $\mathcal{Q}$ serves as the input of the ViT backbone. Specifically, we retain the network structure and weights of SSPEN in the second stage and inject its outputs $\mathcal{A}$ and $\mathcal{B}$ into the pre-trained ViT. The whole process can be formulated as: 
    $$\mathcal{\widehat{A}}, \mathcal{\widehat{B}} = \text{MFIN}(\mathcal{A}, \mathcal{B}, \mathcal{Q}),$$
    where $\text{MFIN}(\cdot,\cdot,\cdot)$ means the Multiscale Feature Interaction Network, denoting the whole pre-trained and fine-tuning network of the second stage, which is illustrated in Fig.~\ref{stage2}; $\mathcal{\widehat{A}}$ and $\mathcal{\widehat{B}}$ represent the results of $\text{MFIN}$ at different scales. Similarly, $\mathcal{\widehat{A}}$ and $\mathcal{\widehat{B}}$ are then fed into the Tail network for decoding to obtain the predicted image of the second stage $\mathcal{O}_{2}$: 
    $$\mathcal{O}_{2} = \text{Tail}_{2}(\mathcal{\widehat{A}}, \mathcal{\widehat{B}}),$$
    where $\text{Tail}_{2}(\cdot,\cdot)$ denotes the Tail network of the second stage.

    \subsection{Local Prior Extraction Stage} \label{sec:stage1}

    Recent works~\cite{wu2021cvt,park2022vision,wang2022pvt} have suggested that convolutions can help Transformers capture the local spatial information in a better way. Inspired by this, we propose to extract local priors from the pre-trained CNN models and inject them to the ViT blocks. The Local Prior Extraction stage network is shown in Fig.~\ref{stage1}. To expand the applicability of our method, we aim to fine-tune a simple pre-trained network, without complicated scale changes and dense skip connections. Therefore, we choose the EDSR~\cite{lim2017enhanced} network as the pre-trained model which is trained on the DIV2K dataset~\cite{agustsson2017ntire} for the image super-resolution task. 

    In order to keep the prior information at different scales, we independently fine-tune two pre-trained CNNs to obtain spectral features and spatial features from $\mathcal{M}$ and $\mathcal{Q}$, respectively. Note that the names of the features are literally for distinguishing scales. In detail, each CNN layer uses a simple Local Prior Extraction (LPE) block  to extract features with a lower channel dimension, whose structure is shown in Fig.~\ref{stage1}. Therefore, taking the spectral branches as an example, the processing for the $i$-th layer's intermediate feature can be formulated as follows: 

    $$ \mathcal{\widehat{C}}^i_{spe}=\mathrm{LPE}_{spe}^i(\mathcal{C}_{spe}^i),$$
    where $\mathcal{C}_{spe}^i$ denotes the $i$-th layer's intermediate spectral feature, $\mathrm{LPE}_{spe}^i(\cdot)$ denotes the LPE block for processing the $i$-th CNN layer and $\mathcal{\widehat{C}}^i_{spe}\in\mathbb{R}^{H\times W\times d'}$ is the extracted intermediate spectral feature with a reduced channel dimension $d'$. In this way, we obtain $\{\widehat{C}^{1}_{spe},\widehat{C}^{2}_{spe},\cdots,\widehat{C}^{n}_{spe}\}$, which will be concatecated and down projected by a linear layer: 
    $$\mathcal{A}=\text{Proj}_{\downarrow}(\text{Concat}(\widehat{C}^{1}_{spe},\widehat{C}^{2}_{spe},\cdots,\widehat{C}^{n}_{spe})),$$
    where $n$ is the number of CNN layers and $\text{Proj}_{\downarrow}(\cdot)$ denotes the down projection layer. The same process is applied to $\mathcal{C}^i_{spa}$ to obtain $\mathcal{B}$. All of the above structures are collectively referred to as the Spatial-Spectral Priors Extraction Network (SSPEN), whose structure and weights will be directly transferred to the second stage. The lightweight design of SSPEN allows us to extract priors from different network depths without fine-tuning the entire network.

    In the Tail network, we apply Implicit Neural Representation (INR) paradigm~\cite{chen2021learning, tang2021joint, deng2023implicit} to perform multi-scale feature fusion and upsampling. To capture high-frequency components in the network, we employ SIREN~\cite{sitzmann2020implicit} as the activation function in the INR. Following the common training strategy, the upsampled LRMS image $\mathcal{M}_\uparrow$ is directly added to the INR output to obtain the final prediction $\mathcal{O}_{1}$ of this stage, with the complete formula given as: 

    $$
    \mathcal{O}_{1} = \text{INR}(\mathcal{A},\mathcal{B}) + \mathcal{M}_\uparrow,
    $$ where $\text{INR}(\cdot,\cdot)$ denotes the INR interpolation framework.

    \subsection{Multiscale Feature Interaction Stage} \label{sec:stage2}

    The network structure for the multi-scale feature interaction stage is shown in Fig.~\ref{stage2}. In this stage, the network can be roughly divided into two parts: the ViT backbone and the cascaded adapters. To exploit the potential of plain ViT for downstream tasks, we choose the Image Processing Transformer (IPT)~\cite{chen2021pre} as our ViT backbone.
    
    The second stage network consists of cascaded identical adapters. $\mathcal{Q}$ serves as the input to the ViT backbone, while $\mathcal{A}$ and $\mathcal{B}$ serve as the inputs to the cascaded adapters. Note that due to the different scales of the inputs, the patch sizes are $4\times 4$ for $\mathcal{Q}$ and $\mathcal{B}$ and $1\times 1$ for $\mathcal{A}$. As in previous works~\cite{houlsby2019parameter,sung2022lst}, we only fine-tune a smaller intrinsic dimension $k$. Therefore, $\mathcal{A}$ and $\mathcal{B}$ will be linearly projected to $k$ dimensions as inputs to the first layer of adapter, which will be formulated as (taking $\mathcal{A}$ as an example): 
    $$\mathcal{F}^0_{spe}=\text{Proj}_{\downarrow}(\text{Reshape}(\mathcal{A})),$$
    where $\text{Reshape}(\cdot)$ denotes the reshape operation,  
    $\mathcal{F}^0_{spe}$ represents the input of the first adapter. 
    
    For the $j$-th cascaded adapter, apart from the two outputs $\mathcal{F}^j_{spe}$ and $\mathcal{F}^j_{spe}$ of the previous adapter serving as inputs, the output of the $j$-th ViT blocks, \emph{i.e.}, $\mathcal{F}^j_{vit}$, also serves as inputs. Note that a linear layer is applied in advance to obtain $\mathcal{F}^j_{vit}$, for projecting the original intermediate feature of ViT to the feature dimension $k$. The $j$-th adapter also has three outputs, $\mathcal{F}^{j+1}_{spe}$ and $\mathcal{F}^{j+1}_{spa}$, the inputs of the next layer of adapters, and $\mathcal{\widehat{F}}^j_{vit}$, which will be injected back into the ViT backbone with a linear layer projecting into original channel dimension. As is shown in Fig.~\ref{stage2}, each adapter contains three modules, including two Cascade Token Fusioner (CTF) modules and one Cascade Token Injector (CTI) module. The two CTF modules can be separately formulated as: 
    $$\mathcal{F}^{j+1}_{spe} = \text{CTF}^j_{spe}(\mathcal{F}^j_{spe},\mathcal{F}^j_{spa},\mathcal{F}^j_{vit}),$$
    and
    $$\mathcal{F}^{j+1}_{spa} = \text{CTF}^j_{spa}(\mathcal{F}^j_{spe},\mathcal{F}^j_{spa},\mathcal{F}^j_{vit}),$$
    where $\text{CTF}^j_{spe}(\cdot,\cdot,\cdot)$ is the CTF module to obtain $\mathcal{F}^{j+1}_{spe}$ and similar to $\text{CTF}^j_{spa}$. The CTI module can be formulated as: 
    $$\mathcal{\widehat{F}}^j_{vit} = \text{CTI}^j(\mathcal{F}^{j+1}_{spe},\mathcal{F}^{j+1}_{spa},\mathcal{F}^j_{vit}),$$
    where $\text{CTI}^j_{spe}(\cdot,\cdot,\cdot)$ denotes the CTI module.
    
    The output of the last adapter layer is reshaped and fed into the Tail network for decoding. Besides, to avoid an excessively large number of parameters for fine-tuning, tokens are fed into adapters every $t$ layers. Balancing performance and parameter count, $t$ is ultimately set to 4. An ablation study on $t$ is provided in the \emph{Suppl. Mat.}

    \subsubsection{Cascade Token Fusioner Module.}
    As shown in Fig.~\ref{stage2}, CTF module is used to fuse the multi-scale features from two branches and interact with features from ViT. In each adapter, there are two parallel CTF modules, one for processing $\mathcal{F}^j_{spa}$ and the other for $\mathcal{F}^j_{spe}$. Taking the module for processing $\mathcal{F}^j_{spa}$ as an example, $\mathcal{F}^j_{spe}$ goes through a proposed weighting network $w_j(\cdot)$, and the result is element-wise multiplied with $\mathcal{F}^j_{spa}$, which can be formulated as: 
    $$\mathcal{\widetilde{F}}^j_{spa}=\mathcal{F}^j_{spa}\cdot w_j(\mathcal{F}^j_{spe})+\mathcal{F}^j_{spa},
    $$ where $\mathcal{\widetilde{F}}^j_{spa}$ serves as the Key and Value and input into a multi-head cross-attention layer. Meanwhile, $\mathcal{F}^j_{vit}$ serves as the Query of the multi-head cross-attention. The complete process can be written as: 
    
    $$ \mathcal{\widehat{F}}^j_{spa} = \text{Attention}(\mathcal{\widetilde{F}}^j_{spa},\mathcal{F}^j_{vit}) + \mathcal{\widetilde{F}}^j_{spa},
    $$ where $\text{Attention}(\cdot)$ denotes the multi-head cross-attention layer with 4 heads, and the specific structure of $w_j(\cdot)$ is shown in Fig.~\ref{stage2}. Subsequently, $\mathcal{\widehat{F}}^j_{spa}$ is fed into a feed-forward network consisting of a linear layer, a GELU activation function, and another linear layer. Inspired by the Adapter structure in NLP~\cite{houlsby2019parameter}, the intermediate dimension of two linear layers is set relatively low:
    
    $$ \mathcal{F}^{j+1}_{spa} = \mathcal{\widehat{F}}^j_{spa} + \text{FFN}(\mathcal{\widehat{F}}^j_{spa}),
    $$ where $\text{FFN}(\cdot)$ denotes the feed-forward network. The obtained $\mathcal{F}^{j+1}_{spa}$ will be used as the input for next adapter layer.
\begin{table*}[!t] 
		\setlength{\tabcolsep}{1mm}
		\centering
		\small
        \begin{tabular}{@{}cccccccc@{}}
            \toprule
            \multirow{2}{*}{\textbf{Methods}} & \multicolumn{4}{c}{\textbf{Reduced-Resolution}} & \multicolumn{3}{c}{\textbf{Full-Resolution}} \\ \cmidrule(l){2-8} 
            &PSNR &Q8 &SAM &ERGAS
            &$D_\lambda$ &$D_S$ &QNR\\ \midrule

                PanNet         
                &  37.346 $\pm$ 2.688                   &  0.891 $\pm$ 0.093  
                                &  3.613 $\pm$ 0.766                                &\multicolumn{1}{c|}{\makecell[c]{  2.664 $\pm$ 0.688}}  
                                &  \textcolor{red}{\textbf{0.0165 $\pm$ 0.0074}}    & 0.0470 $\pm$ 0.0210  
                                &  0.9374 $\pm$ 0.0271  \\
                
                FusionNet        
                &  38.047 $\pm$ 2.589                               &  0.904 $\pm$ 0.090  
                                &  3.324 $\pm$ 0.698                                &\multicolumn{1}{c|}{\makecell[c]{  2.465 $\pm$ 0.644}}  
                                &  0.0239 $\pm$ 0.0090                              &  0.0364 $\pm$ 0.0137  
                                &  0.9406 $\pm$ 0.0197  \\
                % MUCNN~\cite{wang2021ssconv}             &  38.262 $\pm$ 2.703                               &  0.911 $\pm$ 0.089  
                %                 &  3.206 $\pm$ 0.681                                &\multicolumn{1}{c|}{\makecell[c]{  2.400 $\pm$ 0.617}}  
                %                 &  0.0258 $\pm$ 0.0111                              &  0.0327 $\pm$ 0.0140
                %                 &  0.9424 $\pm$ 0.0205  \\
                
                LAGConv
                & 38.592 $\pm$ 2.778 & 0.910 $\pm$ 0.091
                                & 3.103 $\pm$ 0.558 & \multicolumn{1}{c|}{\makecell[c]{2.292 $\pm$ 0.607}}
                                & 0.0368 $\pm$ 0.0148 & 0.0418 $\pm$ 0.0152
                                & 0.9230 $\pm$ 0.0247 \\
                PMACNet        
                &  38.595 $\pm$ 2.882    &  0.912 $\pm$ 0.092
                                &  3.073 $\pm$ 0.623  &\multicolumn{1}{c|}{\makecell[c]{2.293 $\pm$ 0.532}}
                                &  0.0540 $\pm$ 0.0232                              &  0.0336 $\pm$ 0.0115  
                                &  0.9143 $\pm$ 0.0281  \\
                BiMPan          
                &  38.671 $\pm$ 2.732    &  0.915 $\pm$ 0.087
                                &  2.984 $\pm$ 0.601  &\multicolumn{1}{c|}{\makecell[c]{2.257 $\pm$ 0.552}}
                                &  \textcolor{blue}{\textbf{0.0171 $\pm$ 0.0128}}                             &  0.0334 $\pm$ 0.0144  
                                &  0.9493 $\pm$ 0.0255  \\ 
                HMPNet         
                &  38.684 $\pm$ 2.572    
                                &   0.916 $\pm$ 0.087
                                &  3.063 $\pm$ 0.577  &\multicolumn{1}{c|}{\makecell[c]{2.229 $\pm$ 0.545}}
                                &  0.0183 $\pm$ 0.0077
                                &  0.0532 $\pm$ 0.0060  
                                &  0.9293 $\pm$ 0.0144  \\ 

                PanDiff          
                                &  38.424 $\pm$ 2.686    
                                &  0.898 $\pm$ 0.088
                                &  3.297 $\pm$ 0.601  &\multicolumn{1}{c|}{\makecell[c]{2.467 $\pm$ 0.584}}
                                &  0.0276 $\pm$ 0.0120
                                &  0.0541 $\pm$ 0.0266  
                                &  0.9201 $\pm$ 0.0364  \\ 
                PanMamba           
                                &  38.602 $\pm$ 2.788    
                                &  0.916 $\pm$ 0.090
                                &  2.940 $\pm$ 0.540  &\multicolumn{1}{c|}{\makecell[c]{2.240 $\pm$ 0.510}}
                                &  0.0203 $\pm$ 0.0071
                                &  0.0422 $\pm$ 0.0141  
                                &  0.9395 $\pm$ 0.0201  \\ 
                CANNet         
                & \textcolor{blue}{\textbf{38.908 $\pm$ 2.749}}     &  \textcolor{blue}{\textbf{0.920 $\pm$ 0.084}}
                                &  \textcolor{blue}{\textbf{2.930 $\pm$ 0.593}}     &\multicolumn{1}{c|}{\makecell[c]{\textcolor{blue}{\textbf{2.158 $\pm$ 0.515}}}}  
                                &  0.0196 $\pm$ 0.0083                             &  \textcolor{red}{\textbf{0.0301 $\pm$ 0.0074}}  
                                &  \textcolor{blue}{\textbf{0.9510 $\pm$ 0.0132}}  \\
                Ours                             &  \textcolor{red}{\textbf{39.473 $\pm$ 2.626}}     &  \textcolor{red}{\textbf{0.923 $\pm$ 0.081}}  
                                &  \textcolor{red}{\textbf{2.917 $\pm$ 0.560}}      &\multicolumn{1}{c|}{\makecell[c]{  \textcolor{red}{\textbf{2.149 $\pm$ 0.492}}}}  
                                &  0.0173 $\pm$ 0.0076   &  \textcolor{blue}{\textbf{0.0304 $\pm$ 0.0086}}
                                &  \textcolor{red}{\textbf{0.9538 $\pm$ 0.0146}}  \\
              
            \midrule
            \textbf{Ideal value} &$\mathbf{+\infty}$ &\textbf{1} &\textbf{0} &\textbf{0} &\textbf{0} &\textbf{0} &\textbf{1}\\ 
            \bottomrule				
        \end{tabular}
        \caption{The average and standard deviation calculated for all the compared approaches on 20 reduced-resolution and 20 full-resolution samples of WV3 dataset. Best results are in \textcolor{red}{Red}, second-best in \textcolor{blue}{Blue}.}
        \label{wv3tab}
	\end{table*}
    \begin{figure*}[t]
        \scriptsize
        \setlength{\tabcolsep}{0.7pt}
        \centering
        \includegraphics[width=\textwidth]{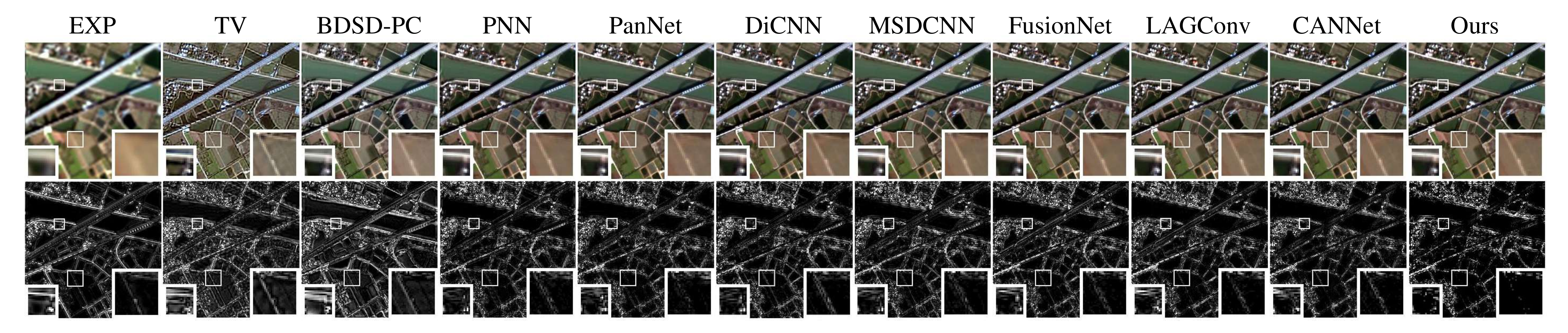}
        \caption{Qualitative evaluation result comparisons with previous pansharpening methods on GF2 reduced-resolution dataset. The first row consists of natural color output, while the second row presents the absolute error maps.}
        \label{gf2}
    \end{figure*}
    \begin{figure*}[t]
        \scriptsize
        \setlength{\tabcolsep}{0.7pt}
        \centering
        \includegraphics[width=\textwidth]{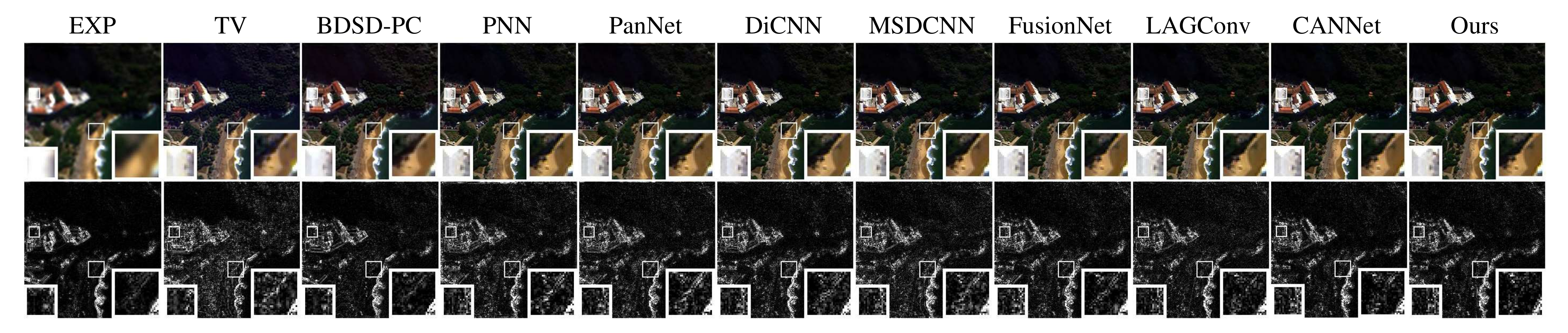}
        \caption{Qualitative evaluation result comparisons with previous pansharpening methods on WV3 reduced-resolution dataset.}
        \label{wv3}
        \vspace{-3pt}
    \end{figure*}
    
    \subsubsection{Cascade Token Injector Module.}
    As shown in Fig.~\ref{stage2}, CTI module is used to inject the spatial and spectral priors from the adapters into the ViT. Specifically, we first concatenate $\mathcal{F}^{j+1}_{spa}$ and $\mathcal{F}^{j+1}_{spe}$ along the feature dimension and project the feature dimension to $k$ using a linear layer, yielding $\mathcal{F}^{j+1}_{fus}$ as the Query for the cross-attention:
    
    $$ \mathcal{F}^{j+1}_{fus} = \text{Linear}(\text{Concat}(\mathcal{F}^{j+1}_{spa},\mathcal{F}^{j+1}_{spe})). $$
    
    \noindent At the same time, the intermediate features from the $i$-th layer ViT, \emph{i.e.}, $\mathcal{F}^j_{vit}$, serve as the Key and Value of the cross-attention and the whole process can be formulated as:
    
    $$ \mathcal{\widehat{F}}_{vit}^j = s_j\cdot\text{Attention}(\mathcal{F}^j_{vit},\mathcal{F}^{j+1}_{fus})+\mathcal{F}^j_{vit},
    $$ where $s_j \in \mathbb{R}$ is a trainable parameter initialized to 0. The obtained $\mathcal{\widehat{F}}_{vit}^j$ is injected into the ViT backbone.

        \begin{table}[ht]
            \small
            \begin{tabular}{l|ccc}
            \toprule
            \textbf{Method} & SAM & ERGAS & Q4 \\ \midrule
            % EXP & 1.820$\pm$0.403 & 2.366$\pm$0.554 & 0.812$\pm$0.051 \\
            % TV & 1.918$\pm$0.398 & 1.745$\pm$0.405 & 0.905$\pm$0.027 \\
            % MTF-GLP-FS & 1.655$\pm$0.385 & 1.589$\pm$0.395 & 0.897$\pm$0.035 \\
            % BDSD-PC & 1.681$\pm$0.360 & 1.667$\pm$0.445 & 0.892$\pm$0.035 \\
            % PNN & 1.048$\pm$0.226 & 1.057$\pm$0.236 & 0.960$\pm$0.010 \\
            PanNet & 0.997$\pm$0.212 & 0.919$\pm$0.191 & 0.967$\pm$0.010 \\
            % DiCNN & 1.053$\pm$0.231 & 1.081$\pm$0.254 & 0.959$\pm$0.010 \\
            FusionNet & 0.974$\pm$0.212 & 0.988$\pm$0.222 & 0.964$\pm$0.009 \\
            LAGConv & 0.786$\pm$0.148 & 0.687$\pm$0.113 & 0.980$\pm$0.009 \\
            HMPNet & 0.803$\pm$0.144 & \textcolor{red}{\textbf{0.562$\pm$0.107}} & 0.986$\pm$0.005 \\
            PanDiff & 0.892$\pm$0.129 & 0.755$\pm$0.108 & 0.979$\pm$0.011 \\
            PanMamba & \textcolor{blue}{\textbf{0.688$\pm$0.129}} & 0.647$\pm$0.103 & 0.939$\pm$0.022 \\
            CANNet & 0.707$\pm$0.148 & 0.630$\pm$0.128 & \textcolor{red}{\textbf{0.983$\pm$0.006}}\\
            
            Ours & \textcolor{red}{\textbf{0.702$\pm$0.143}} & \textcolor{blue}{\textbf{0.615$\pm$0.105}} & \textcolor{blue}{\textbf{0.981$\pm$0.007}} \\ \midrule
            
            \textbf{Ideal value} & \textbf{0} & \textbf{0} & \textbf{1} \\ \bottomrule
            \end{tabular}
            \caption{Average quantitative metrics and standard deviation on 20 reduced-resolution for the GF2 dataset. Some traditional methods and CNN methods are compared.}
            \label{gf2tab}
        \end{table}
    
        \begin{table}[ht]
            \small
            \begin{tabular}{l|ccc}
            \toprule
            \textbf{Method} & SAM & ERGAS & Q4 \\ \midrule
    
            % EXP & 8.435$\pm$1.925 & 11.819$\pm$1.905 & 0.584$\pm$0.075 \\
            % TV & 7.565$\pm$1.535 & 7.781$\pm$0.699 & 0.820$\pm$0.090 \\
            % MTF-GLP-FS & 7.793$\pm$1.816 & 7.374$\pm$0.724 & 0.835$\pm$0.088 \\
            % BDSD-PC & 8.089$\pm$1.980 & 7.515$\pm$0.800 & 0.831$\pm$0.090 \\
            % PNN & 5.205$\pm$0.963 & 4.472$\pm$0.373 & 0.918$\pm$0.094 \\
            PanNet & 5.791$\pm$1.184 & 5.863$\pm$0.888 & 0.885$\pm$0.092 \\
            % DiCNN & 5.380$\pm$1.027 & 5.135$\pm$0.488 & 0.904$\pm$0.094 \\
            FusionNet & 4.923$\pm$0.908 & 4.159$\pm$0.321 & 0.925$\pm$0.090 \\
            LAGConv & 4.547$\pm$0.830 & 3.826$\pm$0.420 & 0.934$\pm$0.088 \\
            BiMPan  & 4.586$\pm$0.821 & 3.839$\pm$0.319 & 0.931$\pm$0.091 \\
            HMPNet & 4.562$\pm$0.871 & 3.809$\pm$0.415 & 0.933$\pm$0.094 \\
            PanDiff & 4.575$\pm$0.736 & 3.742$\pm$0.310 & 0.935$\pm$0.090 \\
            CANNet & \textcolor{red}{\textbf{4.507$\pm$0.835}} & \textcolor{red}{\textbf{3.652$\pm$0.327}} & \textcolor{blue}{\textbf{0.937$\pm$0.083}}\\
            Ours & \textcolor{blue}{\textbf{4.511$\pm$0.810}} & \textcolor{blue}{\textbf{3.593$\pm$0.356}} & \textcolor{red}{\textbf{0.938$\pm$0.077}} \\ \midrule
            \textbf{Ideal value} & \textbf{0} & \textbf{0} & \textbf{1} \\ \bottomrule
            \end{tabular}
            \caption{Average quantitative metrics and standard deviation on 20 reduced-resolution for the QB dataset. Some traditional methods and CNN methods are compared.}
            \label{qbtab}
        \end{table}    
\section{Experiment}

To validate the performance of our network, we conducte a set of experiments on different pansharpening datasets. Our results surpass the current state-of-the-art methods.
    
    \subsection{Experiment Settings}

        \subsubsection{Datasets.}
        We investigate the effectiveness of the proposed method on a wide range of datasets, including an 8-band dataset from WorldView-3 (WV3) sensor, and 4-band datasets from QuickBird (QB) and GaoFen-2 sensors. Notably, we leverage Wald’s protocol to simulate the source data due to the unavailability of ground truth (GT) images.

        Taking WV3 as an instance, we use 10000 PAN/LRMS/GT image pairs (64 × 64) for network training. For testing, we take 20 PAN/LRMS/GT image pairs (256 × 256) for reduced-resolution evaluation, and 20 PAN/LRMS image pairs (512 × 512) thanks to the absence of GT images on the full-resolution assessment.

        \subsubsection{Benchmarks.}
        To assess the performance of our approach, we qualitatively and quantitatively compare the proposed method with current state-of-the-art pansharpening methods, including PanNet~\cite{yang2017pannet}, FusionNet~\cite{deng2020detail}, LAGConv~\cite{jin2022lagconv}, PMACNet~\cite{liang2022pmacnet}, BiMPan~\cite{hou2023bidomain}, HMPNet~\cite{tian2023interpretable},  PanDiff~\cite{meng2023pandiff}, PanMamba~\cite{he2024pan}, CANNet~\cite{duan2024content}. Noted that all DL-based approaches considered for comparison are trained using the same datasets as our method, while their hyperparameter settings follow the respective original papers.

        \begin{table}[ht]
            \small
            \setlength{\tabcolsep}{1.9mm}
            \begin{tabular}{l|ccc}
            \toprule
            \textbf{Method} & SAM & ERGAS & Param.(M) \\ \midrule
            VPT & - & - & 0.53(0.46$\%$)\\
            LoRA & 4.337$\pm$0.911 & 3.374$\pm$0.709 & 1.3(1.13$\%$) \\
            Adapter & 3.809$\pm$0.844 & 2.793$\pm$0.654 & 2.07(1.79$\%$) \\
            LST & 3.342$\pm$0.667 & 2.507$\pm$0.619& 1.76(1.53$\%$) \\
            Full Fine-tune & \textcolor{blue}{\textbf{3.288$\pm$0.637}} & \textcolor{blue}{\textbf{2.473$\pm$0.662}} & 115.2(100$\%$) \\
            Ours & \textcolor{red}{\textbf{2.917$\pm$0.560}} & \textcolor{red}{\textbf{2.149$\pm$0.492}}& 2.19(1.85$\%$) \\ \midrule
            \textbf{Ideal value} & \textbf{0} & \textbf{0} & - \\ \bottomrule
            \end{tabular}
            \caption{Average quantitative metrics and standard deviation of fine-tuning methods on WV3 dataset.}
            \label{finetune}
            \vspace{-7pt}
        \end{table}

        \begin{table}[ht]
            \setlength{\tabcolsep}{1.5mm}
            \small
            \begin{tabular}{l|cccc}
            \toprule
            \textbf{Method} & SAM & ERGAS & Q8  \\ \midrule
            w/o two-stage & 3.573$\pm$0.682 & 2.576$\pm$0.619 & 0.906$\pm$0.095\\
            Replace INR & 2.957$\pm$0.640 & 2.381$\pm$0.916 & 0.916$\pm$0.090 \\
            w/o CTF module & \textcolor{blue}{\textbf{2.942$\pm$0.611}} & \textcolor{blue}{\textbf{2.155$\pm$0.508}} & \textcolor{blue}{\textbf{0.919$\pm$0.090}}  \\
            w/o CTI module & 3.301$\pm$0.642 & 2.506$\pm$0.649 & 0.907$\pm$0.088 \\
            Ours & \textcolor{red}{\textbf{2.917$\pm$0.560}} & \textcolor{red}{\textbf{2.149$\pm$0.492}} & \textcolor{red}{\textbf{0.923$\pm$0.081}} 
            \\ \bottomrule
            \end{tabular}
            \caption{Ablation study on WV3 dataset.}
            \label{ablationtab}
            \vspace{-7pt}
        \end{table}

        \subsubsection{Evaluation Metrics.}
        To assess the performance of the reduced-resolution dataset, we utilize four evaluation metrics: PSNR, $\text{Q}_{2n}$~\cite{garzelli2009hypercomplex}, SAM~\cite{boardman1993automating}, and ERGAS~\cite{wald2002data}. For the evaluation of the full-resolution dataset, we employ three metrics: $D_\lambda$, $D_s$, and QNR~\cite{vivone2014critical}.

    \begin{figure}[t]
        \scriptsize
        \setlength{\tabcolsep}{0.7pt}
        \centering
        \includegraphics[width=0.42\textwidth]{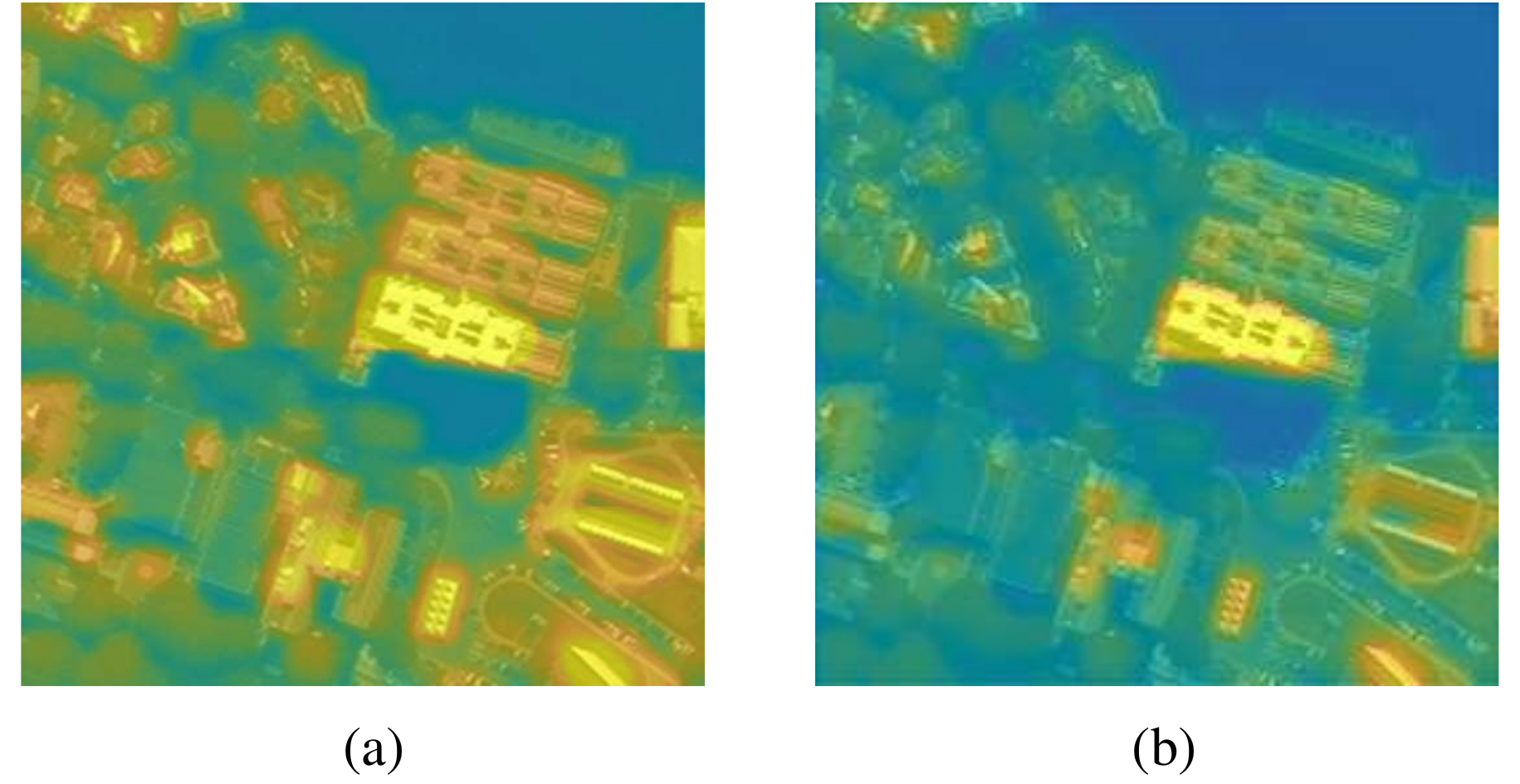}
        \caption{(a) and (b) are heat maps of the intermediate features of the first and second stage networks, respectively. It can be seen that compared to CNN networks, ViT with priors injected is more accurate and efficient in capturing image semantics.}
        \label{feature}
        \vspace{-7pt}
    \end{figure}

    \begin{figure}[ht]
        \scriptsize
        \setlength{\tabcolsep}{0.7pt}
        \centering
        \includegraphics[width=0.46\textwidth]{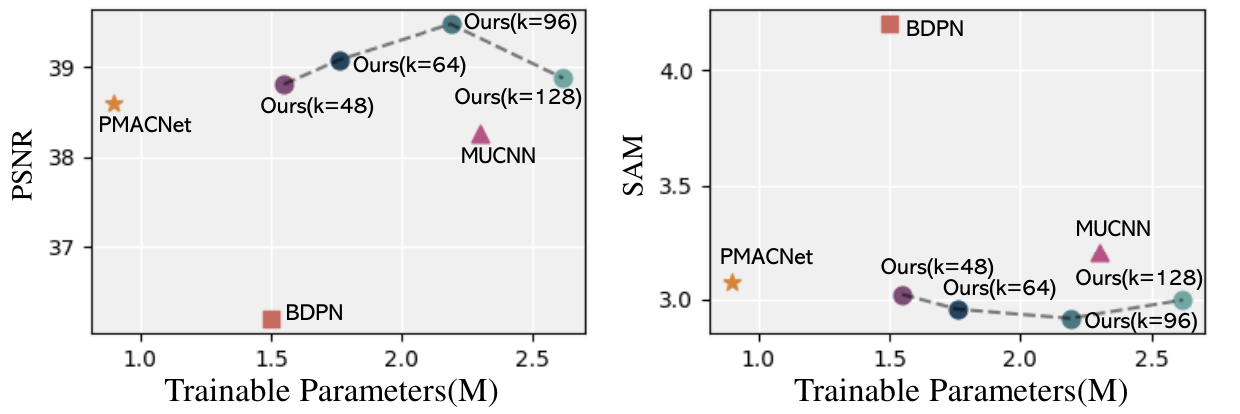}
        \caption{Comparison between SOTA methods and our method with different $k$,  across varying parameters.}
        \label{PSNRSAM}
        \vspace{-7pt}
    \end{figure}
    
    \subsection{Comparison} \label{finetune_exp}
    Tabs.~\ref{wv3tab}, ~\ref{gf2tab}~and ~\ref{qbtab} present a detailed comparison between our method and various state-of-the-art methods on three benchmark datasets: WorldView-3 (WV3), QuickBird (QB), and GaoFen-2 (GF2). The results demonstrate the robustness of PanAdapter across different datasets and its consistent capability to produce high-quality pansharpened images. 

    Tab.~\ref{finetune} compares the performance and parameter count/parameter ratio of other fine-tuning methods with our method on WV3. The specific structure of the Tail network is consistent with our method. It is worth mentioning that all existing fine-tuning methods converge slowly and are very sensitive to hyperparameters due to the lack of two-stage training and the injection of CNN priors.

    \subsection{Visualization}
    Visual comparisons depicted in Fig.~\ref{gf2}~and ~\ref{wv3} reveal that our method generates results that are perceptually closer to the ground truth. It can be seen from the residual images that our method recovers parts such as houses and roads very well, which is most likely due to the information of natural image datasets. Moreover, we visualize the intermediate features of the network completing the first stage of training and completing the second stage of training, respectively, as shown in Fig.~\ref{feature}.Lastly, the performance comparison results, as shown in Fig.~\ref{PSNRSAM}, demonstrate that our method consistently surpasses existing state-of-the-art approaches across various levels of parametric complexity. The adjustment of the intrinsic dimension $k$ serves as a mechanism for modifying the model’s parameter count to achieve these results.

    \subsection{Ablation Study}
        The effectiveness of the two-stage fine-tuning strategy and the multiscale adapters is proven by ablation study on WV3 dataset, as shown in Tab.~\ref{ablationtab}. The experiments, including stacking the network without the two-stage training strategy, replacing INR with a common convolutional upsampling module, and removing CFI and CTI modules, demonstrate that all proposed methods positively impact performance.

\section{Conclusion}

In this paper, we propose an efficient fine-tuning method for pansharpening, \emph{i.e.}, PanAdapter. It successfully fine-tunes pre-trained image restoration models to the pansharpening task through two-stage training and multi-scale feature extraction, thus effectively reducing the difficulty of domain transfer. The dual-branch adapters fuse and inject the spatial priors and spectral priors separately, enhancing feature extraction efficiency. Our PanAdapter outperforms state-of-the-art pansharpening methods on several datasets, providing a new paradigm for related image fusion tasks.

% \section*{Acknowledgements}

\appendix

\bibliography{aaai25}

\end{document}